\documentclass[letterpaper, 10 pt, conference]{ieeeconf}  %
\let\labelindent\relaxComment 

\IEEEoverridecommandlockouts                              

\usepackage{amsmath,amsfonts}
\usepackage{threeparttable}
\usepackage{graphicx}
\usepackage{siunitx}
\usepackage{multirow}
\usepackage{proof}
\usepackage{makecell}
\usepackage{array}
\usepackage{amsmath}
\usepackage{amssymb}
\usepackage{epstopdf}
\usepackage{float}
\usepackage{amssymb}
\usepackage{nomencl}
\usepackage{enumerate}
\usepackage{enumitem}
\usepackage[linesnumbered,ruled,vlined]{algorithm2e}
\usepackage{subfigure}
\usepackage{bm}

\usepackage{amsthm}
\usepackage{multirow}
\usepackage{array}
\usepackage[caption=false,font=normalsize,labelfont=sf,textfont=sf]{subfig}
\usepackage{textcomp}
\usepackage{stfloats}
\usepackage{url}
\usepackage{xcolor} 
\usepackage{verbatim}

\usepackage{graphicx}
\usepackage{cite}
\usepackage{booktabs}
\newcommand{\RNum}[1]{\uppercase\expandafter{\romannumeral #1\relax}}
\usepackage[english]{babel}
\newtheorem{definition}{Definition}
\newtheorem{theorem}{{Theorem}}
\usepackage{wrapfig}
\usepackage{enumitem}
\usepackage[capitalize]{cleveref}
\title{\LARGE \bf
Dynamic Residual Safe Reinforcement Learning for Multi-Agent Safety-Critical Scenarios Decision-Making
}

\author{Kaifeng Wang, Yinsong Chen, Qi Liu, Xueyuan Li*, Xin Gao*
\thanks{This work was supported by National Key R\&D Plan of China (Grant No.2024YFB3411301)}
\thanks{(Corresponding author: Xueyuan Li and Xin Gao)}
\thanks{Kaifeng Wang, Yinsong Chen, Qi Liu, Xueyuan Li, and Xin Gao are with the School of Mechanical Engineering, Beijing Institute of Technology, Beijing, China. (E-mails: 3120230311@bit.edu.cn; 3220240416@bit.edu.cn; 3120195257@bit.edu.cn; lixueyuan@bit.edu.cn; gaoxin2000@bit.edu.cn)}%
}
\begin{document}
\maketitle
\thispagestyle{empty}
\pagestyle{empty}

\begin{abstract}

In multi-agent safety-critical scenarios, traditional autonomous driving frameworks face significant challenges in balancing safety constraints and task performance. These frameworks struggle to quantify dynamic interaction risks in real-time and depend heavily on manual rules, resulting in low computational efficiency and conservative strategies. To address these limitations, we propose a Dynamic Residual Safe Reinforcement Learning (DRS-RL) framework grounded in a safety-enhanced networked Markov decision process. It's the first time that the weak-to-strong theory is introduced into multi-agent decision-making, enabling lightweight dynamic calibration of safety boundaries via a weak-to-strong safety correction paradigm. Based on the multi-agent dynamic conflict zone model, our framework accurately captures spatiotemporal coupling risks among heterogeneous traffic participants and surpasses the static constraints of conventional geometric rules. Moreover, a risk-aware prioritized experience replay mechanism mitigates data distribution bias by mapping risk to sampling probability. Experimental results reveal that the proposed method significantly outperforms traditional RL algorithms in safety, efficiency, and comfort. Specifically, it reduces the collision rate by up to 92.17\%, while the safety model accounts for merely 27\% of the main model's parameters. 
\end{abstract}

\section{INTRODUCTION}

Breakthroughs in artificial intelligence are propelling autonomous driving technology from laboratory validation toward a critical transformation phase for commercialization. In California, several companies, including Waymo and Cruise, have already acquired open-road testing permits. However, the application of intelligent transportation systems remains constrained by safety and trust gaps in open-road environments. Vehicle safety is closely linked to performance in safety-critical scenarios \cite{safety-critical}, where autonomous driving systems must not only address millisecond-level decision-making demands but also maintain failure probabilities at orders of magnitude lower than those of human drivers. The extreme complexity of open-road environments and the behavioral uncertainty of traffic participants \cite{interactive} make the "long-tail problem" increasingly apparent. Addressing safety-critical scenarios has thus emerged as a central bottleneck limiting the commercialization of autonomous driving. 

\begin{figure}[htbp]
    \centering
    \includegraphics[scale=0.3]{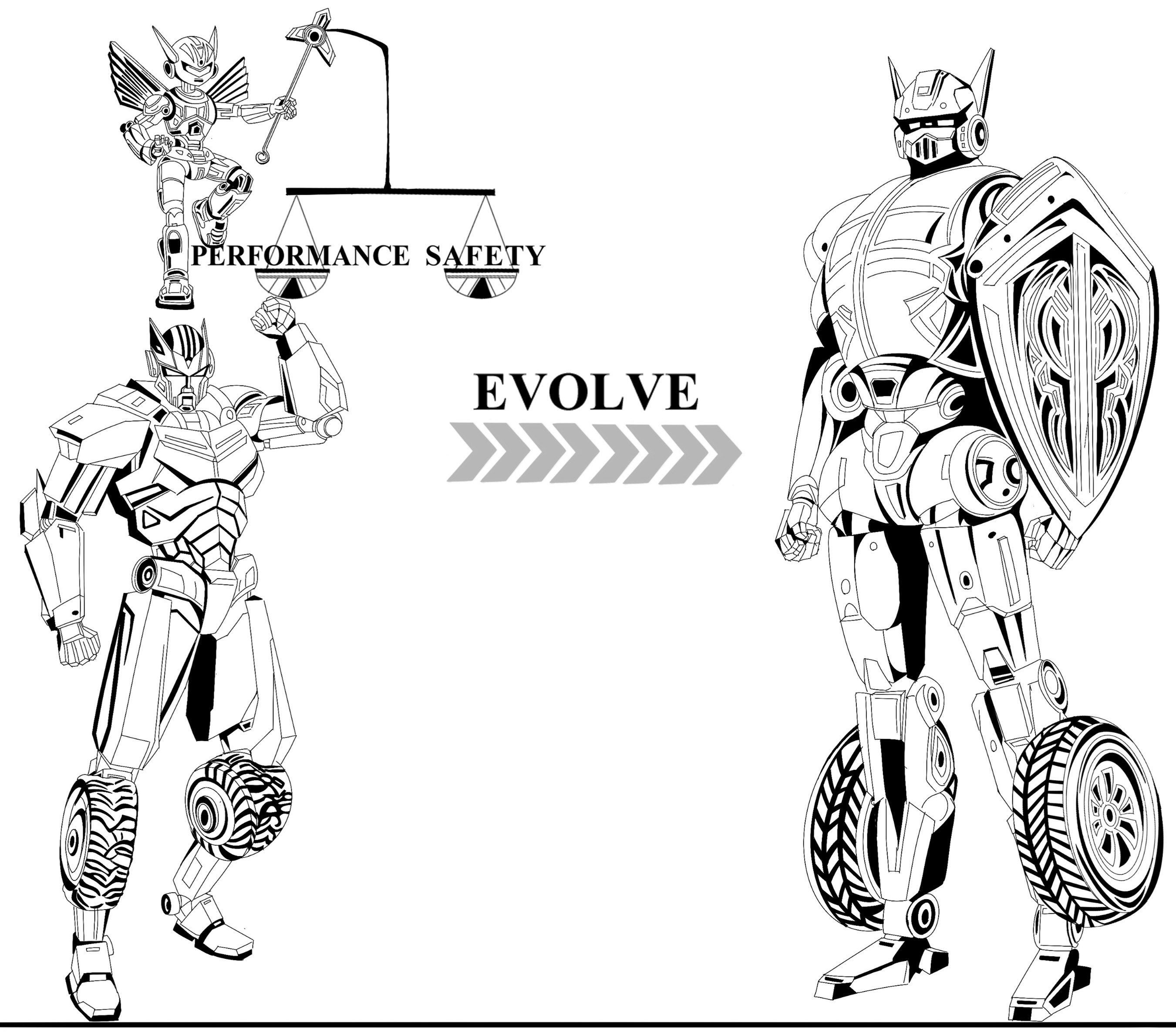} 
    \caption{An illustration of the proposed methodology. Our method is inspired by the weak-to-strong correction  \cite{weak-to-strong-correction} and introduces a lightweight safety model to balance performance and safety. It enables the agent to evolve into a safer and more robust entity while preserving its original performance.}
    \label{fig: inspiration figure}
\end{figure}

Some studies have improved decision-making in single-agent safety-critical scenarios. However, given the complexity of urban road networks, heavy traffic, and diverse participant behaviors, future urban environments will inevitably feature multiple autonomous vehicles (AVs) operating in tandem \cite{future}. This emergent collective intelligence poses significant challenges to rule-based decision-making systems, expanding safety-critical scenarios and introducing greater complexity. Multi-agent safety-critical scenarios (MASCS) are dynamic, high-risk conditions arising from spatiotemporal coupling among heterogeneous road users (e.g., AVs, background vehicles, pedestrians) in shared spaces. These scenarios can be viewed as a multi-agent dynamic conflict zone, characterized by potential high-risk events and chain risk propagation among multiple entities when indicators (e.g., time to collision, post-encroachment time) exceed thresholds. 

Research on MASCS decision-making faces three principal scientific challenges: 

\begin{enumerate}[label=(\alph*)]
\item In safety-critical scenarios, existing approaches frequently introduce extra constraints or parameters to maintain safety, causing parameter expansion and reduced efficiency. It not only leads to computational delays but also imposes overly cautious constraints that limit task performance. 
\item The diverse behaviors of traffic participants introduce competitive and cooperative interactions with highly dynamic conflict characteristics. Traditional geometry-based conflict recognition is inadequate, necessitating conflict zone modeling with dynamic topological relationships to capture these complex interactions.  
\item The long-tail effect results in an oversaturation of routine driving segments and a scarcity of safety-critical ones. This skewed data distribution underestimates decision-making capabilities in safety-critical settings. Thus, effective sampling methods under imbalanced data conditions remain a key challenge. 
\end{enumerate}

To address these challenges, we propose a dynamic residual safe reinforcement learning (DRS-RL) framework and develop a safety-enhanced networked Markov decision process (MDP). We introduce a multi-agent dynamic conflict zone (MADCZ) model that accurately captures the dynamic interactions among traffic participants. Additionally, we design a risk-aware prioritized experience replay (PER) mechanism to enhance decision-making in safety-critical scenarios. Finally, simulation experiments on a comprehensive MASCS set demonstrate the significant advantages of our approach in enhancing both safety and task performance.  The main contributions of this paper are as follows: 

\begin{enumerate}[label=(\alph*)]
\item We propose a safety-enhanced networked MDP and a DRS-RL framework. By leveraging a lightweight model for weak-to-strong safety correction, we effectively balance safety constraints and task performance, thereby substantially enhancing parameter efficiency. 
\item We develop a MADCZ model that accurately captures and quantifies potential risks in complex interactions by leveraging dynamic topological structures and spatiotemporal conflict zone modeling techniques. 
\item We propose a risk-aware PER method that effectively mitigates data distribution bias by mapping risk intensity to sampling probability. 
\end{enumerate}

\section{Related Work}

\subsection{Decision-Making Methods in Safety-critical Scenarios}

Autonomous driving decision-making has made significant progress in conventional driving scenarios, but its shortcomings in safety-critical scenarios are gradually gaining attention from researchers. Niu et al. \cite{niu-dr2l}  proposed a domain randomization RL framework to progressively generate complex corner cases, thereby enabling AVs to achieve enhanced safety under hazardous conditions. Under high-speed cut-in emergency scenarios,  Wang et al.\cite{wang2024safety}  proposed a decision-state machine that employs an oriented bounding box collision detection method and longitudinal prediction distance to effectively assess collision risks.  Fu et al. \cite{fu2020decision}  proposed an emergency braking strategy to address sudden lane changes or abrupt braking by a lead vehicle, achieving an approximate 15\% reduction in collision rates. Li et al.\cite{li2024human}  developed a cumulative information processing method for drivers based on the drift-diffusion model to elucidate decision-making in rear-end collision scenarios. Zhou et al. \cite{zhou2022safety}  investigated safety-critical control strategies within a leader-follower cruise control framework. They utilized a control barrier function (CBF) to ensure safe inter-vehicle distances during emergency speed adjustments. CBF has also been applied to the autonomous safe navigation of robots \cite{CBF1} and unmanned aerial vehicles \cite{CBF2}, \cite{CBF3}. Hu et al. \cite{hu2022rear} examined the severe consequences of rear-end collisions involving hazardous materials transport vehicles, which may lead to explosions. They proposed an actor-critic-based method for collision avoidance. 

Notably, the aforementioned studies predominantly concentrate on safety-critical strategies for single AVs. Although advances have improved the safety of AVs, research on multi-vehicle scenarios remains insufficient. MASCS are characterized by high interactivity, thereby necessitating further in-depth research. Toghi et al. \cite{toghi2021altruistic}  investigated the egoistic driving behavior in ramp merging scenarios. They employed an altruistic maneuver-oriented reward function to enhance merging safety. Li et al. \cite{li2024autonomous} introduced a global sorting-local gaming framework to tackle dense multi-vehicle interaction decision-making at unsignalized intersections. These studies highlight the potential to enhance traffic safety and efficiency. However, further research is required to address the complex challenges in MASCS.

\subsection{Decision-Making Methods based on Safe RL}

Safe RL incorporates safety constraints within the learning framework, which is particularly well-suited for AVs systems with low-risk tolerance. Kamran et al. \cite{kamran2020risk} proposed a risk-aware deep Q-network (DQN) approach for longitudinal decision-making at obstructed intersections. The reward function incorporates risk-aware incentives, achieving a success rate exceeding 80\%. Xu et al.\cite{xu2020decision} predict the trajectories of vehicles based on the constant turn rate and acceleration model to ensure safe driving in lane reduction scenarios. However, it relies on manually crafted safety rules, limiting adaptability.  Hanna et al. \cite{krasowski2022safe}  introduced a Safe RL approach, incorporating a safety layer based on invariably safe braking sets. However, the safety constraints led to a lower goal-reaching rate in certain datasets compared to the baseline.  Li et al. \cite{li2023safe}  facilitated safe lane-changing exploration by developing a safety detection model.  The proposed method significantly reduced collision rates, but it led to a slight reduction in average speed. Luo et al.\cite{luo2024platoon} proposed a Lyapunov-based soft actor-critic (SAC) algorithm that formulates a constrained MDP. Vehicle Platoon control was also studied in \cite{luo2024safe}, where a risk probability prediction-based SAC method was introduced.  However, the inclusion of a risk prediction model with a deeper architecture than the policy model results in increased model complexity, potentially introducing latency issues. 

Existing methods face several limitations: (a) Rigid safety constraints degrade task performance. (b) Manually crafted rules reduce algorithm adaptability. (c) Complex verification models impose a computational burden.

\section{Methodology}

\begin{figure}[bt!]
    \centering
    \includegraphics[width=1\linewidth]{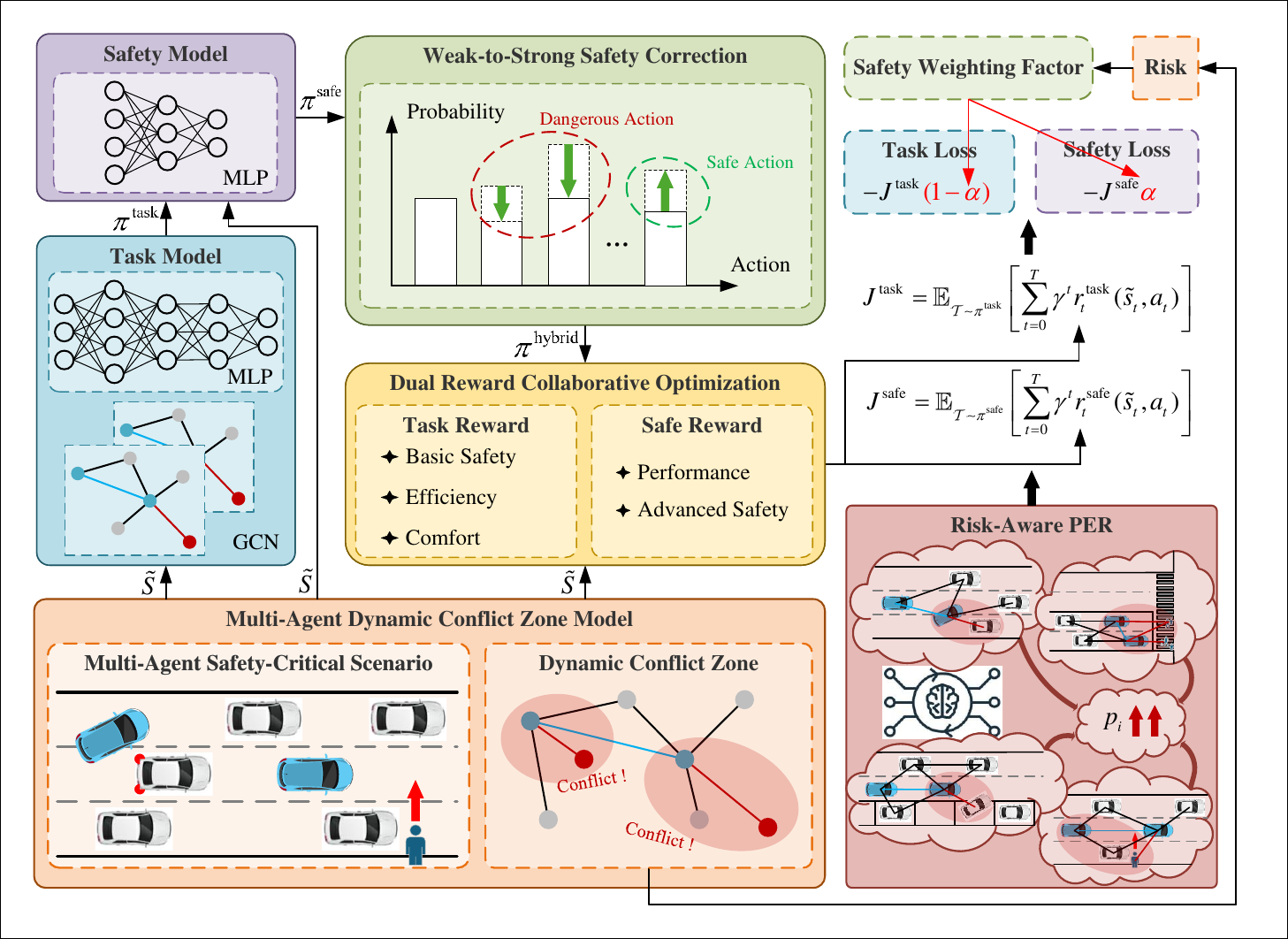}
    \caption{Overall architecture of the proposed method. Multi-agent safety-critical scenarios are modeled as a dynamic conflict zone. Based on this representation, the DRS-RL algorithm generates hybrid strategies through the weak-to-strong safety correction paradigm. The task and safety models are optimized through the risk-aware PER method and dual-reward collaborative optimization, emphasizing the learning for safety-critical segments.}
    \label{fig: DRS-RL}
\end{figure}

This section provides a systematic overview of our core methodology. The safety-enhanced networked MDP model and the DRS-RL framework are proposed. It achieves residual correction through a lightweight safety model. Additionally, a MADCZ modeling method is designed to quantify interactive threats based on dynamic topology and spatiotemporal coupling risks. The overall architecture of the proposed method is illustrated in \cref{fig: DRS-RL}.


\subsection{Safety-Enhanced Networked MDP}

The traditional MDP framework suffers from two fundamental limitations: (a) Its single-agent assumption is insufficient for modeling multi-agent collaborative decision-making, and (b) the assumption of fully observable global states is infeasible in complex, dynamic environments. To address these limitations, we extend the networked MDP \cite{ma2024efficient} by developing a safety-enhanced variant. It employs incremental action corrections to prevent unsafe action space exploration. 

\begin{definition}[Safety-Enhanced Networked MDP]
A safety-enhanced networked MDP is formally defined as the tuple $N = (\tilde S,{A^{{\rm{task}}}},{A^{{\rm{safe}}}},A,\alpha ,P,{R^{{\rm{task}}}},{R^{{\rm{safe}}}},\gamma )$, where 
$\tilde{S}$ is the fused global state space, which is formed by merging the local observations of the agents;
${A^{{\rm{task}}}}$ is the task action space, generated by the task policy model ${\pi ^{{\rm{task}}}}$;
${A^{{\rm{safe}}}}$ is the safety action space, generated by the safety policy model ${\pi ^{{\rm{safe}}}}$, its inputs are the state $\tilde S$ and the task policy ${\pi ^{{\rm{task}}}}$;
$A$ is the final action space dynamically synthesized through the residual connection mechanism, calculated as $A = {A^{{\rm{task}}}} + \alpha ({A^{{\rm{safe}}}} - {A^{{\rm{task}}}})$, where $\alpha $ is the safety weighting factor, dynamically adjusted by the real-time risk quantification function;
$P:\tilde S \times A \times \tilde S \to [0,1]$ is the state transition probability function; 
${R^{{\rm{task}}}}$ and ${R^{{\rm{safe}}}}$ are the task and safety reward functions, representing basic performance objectives and risk-avoidance capabilities, respectively;
$\gamma  \in [0,1)$ is the discount factor. 
The task policy and safety policy are updated by maximizing their respective objective functions
\begin{align}
\left\{\begin{array}{l}
J^{\mathrm{task}}=\mathbb{E}_{\mathcal{T} \sim \pi^{\mathrm{task}}}\left[\sum_{t=0}^T \gamma^t r_t^{\mathrm{task}}\left(s_t, a_t\right)\right] \\
J^{\mathrm{safe}}=\mathbb{E}_{\mathcal{T} \sim \pi^{\mathrm{safe}}}\left[\sum_{t=0}^T \gamma^t r_t^{\mathrm{safe}}\left(s_t, a_t\right)\right]
\end{array}\right..
\end{align}
\end{definition}

\begin{figure}[htbp]
    \centering
    \includegraphics[width=\columnwidth]{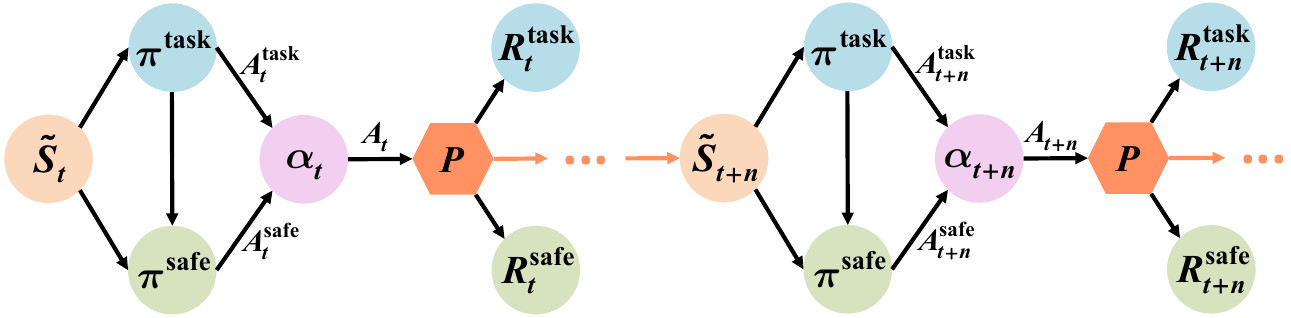}
    \caption{Safety-Enhanced Networked Markov Decision Process.}
    \label{fig: MDP}
\end{figure}

The safety-enhanced networked MDP chain is illustrated in \cref{fig: MDP}. At time step $t$, the environment’s fused global state ${\tilde S_t}$ is fed into the task policy model $\pi^{\rm{task}}$, which outputs the basic action $A_t^{\rm{task}}$ that satisfies the driving objectives. Subsequently, the safety policy model $\pi^{\rm{safe}}$  generates the safety action $A_t^{\rm{safe}}$ based on the current state ${\tilde S_t}$ and the output of the task policy $\pi^{\rm{task}}$. The final action ${A_t}$ is then synthesized through the residual connection mechanism using the safety weighting factor ${\alpha _t}$, and fed into the state transition probability function $P$ to update the state. Finally,  the rewards $R_t^{\rm{task}}$  and $R_t^{\rm{safe}}$ are calculated using the task and safety reward functions, respectively, and are employed to update the corresponding policy networks. 

\subsection{Multi-Agent Dynamic Conflict Zone Model}

In the MADCZ model, the state space encapsulates essential traffic participant information, serving as the foundation for constructing a dynamic conflict zone (DCZ). The DCZ quantifies real-time interactive threats among traffic participants, thereby providing a basis for risk assessment in action-space decision-making. 

\subsubsection{State Space}
In safety-critical scenarios, traffic participants comprise AVs, background vehicles (BVs), and pedestrians (Peds). Their interactions are characterized by significant dynamism and heterogeneity. To address these challenges, we propose a MADCZ modeling approach. By constructing dynamic topological structures and spatiotemporal conflict zones, the model attains precise conflict identification and delivers interpretable decision support.  First, a joint state space is established, defined as
\begin{align}
S = \mathcal{S}^{\text{AVs}} \times \mathcal{S}^{\text{BVs}} \times \mathcal{S}^{\text{Peds}} \times \mathcal{S}^{\text{Road}},
\end{align}
where ${\mathcal{S}^{{\text{AVs}}}}$, ${\mathcal{S}^{{\text{BVs}}}}$, ${\mathcal{S}^{{\text{Peds}}}}$ and ${\mathcal{S}^{{\text{Road}}}}$ represent the state subspaces of AVs, BVs, Peds, and road network, respectively. Each subspace is specifically defined as
\begin{align}
&\left\{
\begin{array}{l}
\mathcal{S}^{\text{Vehs}} = [x, y, \theta, v, l, c, p] \in \mathbb{R}^{22} \\
\mathcal{S}^{\text{Peds}} = [x, y, \theta, v, l, c] \in \mathbb{R}^{10} \\
\mathcal{S}^{\text{Road}} = \left\{ \left(G(V, E) \mid V \in \mathbb{R}^{n \times 22}, E \in \{0,1\}^{n \times n}\right\} \right.
\end{array}
\right.,
\end{align}
where $x$ and $y$ denote the horizontal and vertical coordinates of the traffic participants, $\theta \in [0,360^\circ)$ is the heading angle, $v$ represents the longitudinal velocity,  $l$ and $c$ represent the lane position and traffic participant type, respectively, each encoded as a three-dimensional one-hot vector. $G$ represents the road network topology, where each traffic participant is modeled as a node ${v_i} \in \mathbf{V}$, and $\mathbf{E}$ represents the connections among participants, representing sensor perception or vehicle-to-vehicle (V2V) communication relationships. Additionally, for vehicles, $p$ denotes the relative motion information with respect to surrounding vehicles, defined as
\begin{align}
p = \left[ \Delta d_j, \Delta v_j \right], j = \{f, r, lf, lr, rf, rr \},
\end{align}
where $\Delta {d_j}$ and $\Delta {v_j}$ denote the relative longitudinal distance and the relative velocity between vehicles, and $f$, $r$, $lf$, $lr$, $rf$, $rr$ represent the neighboring vehicles at the front, rear, left front, left rear, right front, and right rear, respectively. If no neighboring vehicle is detected in a given direction, the relative longitudinal distance is assigned the maximum perception range and the relative velocity is set to zero.

\subsubsection{Dynamic Conflict Zone}
In MASCS, we innovatively propose a MADCZ model that quantifies real-time interaction threats among different traffic entities. By incorporating an evolving conflict zone mechanism with an adaptive topological structure, the model precisely characterizes spatiotemporal interactions among heterogeneous traffic participants. The proposed DCZ model is formally defined as
\begin{align}
\Omega_{\mathrm{DCZ}}=\bigcup_{(i, j, k) \in \mathcal{I}}\left\{(x, y, t) \left\lvert\,\left\{\begin{array}{l}
\mathrm{TTC}_{i j}(t) \leq \tau^{\mathrm{V} 2 \mathrm{V}} \\
\operatorname{PET}_{i k}(t) \leq \tau^{\mathrm{V} 2 \mathrm{P}} \\
\exists e(t) \in \mathcal{E}
\end{array}\right\}\right.\right.,
\end{align}
where ${\Omega _{{\text{DCZ}}}} \subseteq {\mathbb{R}^2} \times {\mathbb{R}^+ }$ represents the DCZ, $\mathcal{I} = \{ (i,j,k)|i \in \text{AVs},j \in \text{BVs},k \in \text{Peds}\} $ is the interaction index set; $\tau^{\text{V2V}}$ and $\tau^{\text{V2P}}$ are time to collision (TTC) and post-encroachment time (PET) thresholds, respectively,  $\mathcal{E}$ is the set of typical hazardous events, including leading vehicle emergency braking, pedestrians crossing, and vehicles cutting-in.

Leveraging these spatiotemporal coupling constraints, the model captures potential risks in real-time.  It effectively mitigates risk misjudgments arising from reliance on single indicators and delays inherent in static assumptions, thereby delivering high-precision risk quantification for decision-making.

\subsubsection{Action Space}
The action space represents the joint control commands for AVs, in the form of
\begin{align}
\mathcal{A} = {\mathcal{A}_1} \times {\mathcal{A}_2} \times  \cdots  \times {\mathcal{A}_N} \subset {\mathbb{R}^{2N}},
\end{align}
each AV's individual action space is given by
\begin{align}
{\mathcal{A}_i} = [{a_i},{\delta _i}],{a_i} \in [ - {a_{\text{max}}},{a_{\text{max}}}],{\delta _i} \in [ - {\delta _\text{max}},{\delta _{\text{max}}}],
\end{align}
where $a$ and $\delta$ represent acceleration and steering angle, respectively.

To satisfy the demands for dynamic response accuracy and computational efficiency in safety-critical scenarios, acceleration and steering angle are uniformly discretized into 11 and 13 levels, respectively. Additionally, action coupling constraints are imposed to enhance training efficiency, with the steering control activated exclusively when acceleration is zero. This approach reduces the action space dimensionality from 143 to 23, thereby enhancing algorithm convergence efficiency while maintaining steering stability. 

To mitigate hazardous behaviors during RL exploration, an action space constraint mechanism is developed. These constraints restrict illegal or high-risk actions, ensuring the learning process remains within a safe operational envelope. Specifically, the constraints encompass: (a) preventing off-road excursions; (b) prohibiting reversing maneuvers; (c) limiting U-turns; (d) curbing excessive lane changes; and (e) restricting acceleration when TTC falls below a predefined threshold.

\subsection{Dynamic Residual Safe Reinforcement Learning}

In MASCS, traditional RL algorithms struggle to balance task performance and safety constraints. To address this, as illustrated in \cref{fig: DRS-RL}, we propose the DRS-RL framework that incorporates a safety model to dynamically balance performance and safety.

DRS-RL is built upon a weak-to-strong safety correction paradigm and utilizes a dual-model parallel decision-making framework. Unlike existing approaches that depend on complex safety-layer architectures, it integrates a lightweight safety model comprising only 27\% of the task model’s parameters to perform safety corrections. Specifically, the task policy $\pi^{{\text{task}}}$ generates fundamental actions to achieve driving objectives,  whereas the safety policy $\pi^{{\text{safe}}}$ produces corrective actions to avert collision risks.  These policies are subsequently fused via a dynamic residual connection mechanism, thereby ensuring that overall system performance remains at least equivalent to that of the original task policy. 

\begin{definition}[Hybrid Policy]
The hybrid policy of AVs is defined as follows
\begin{equation}
\begin{aligned}
\pi^{\mathrm{hybrid}}(a \mid s) = \pi^{\mathrm{task}}(a \mid s) + \alpha(s) \Big( \pi^{\mathrm{safe}}(a \mid s, \pi^{\mathrm{task}}) \\\ - \pi^{\mathrm{task}}(a \mid s) \Big),
\end{aligned}
\end{equation}
where $\alpha(s)$ is the safety weighting factor. It is dynamically adjusted according to a real-time risk quantification function,
\begin{align}
\alpha(s) = f(\mathrm{risk}(s)) = 
\begin{cases} 
\alpha_0, & \textup{if } \mathrm{risk}(s) \geq \tau^{\mathrm{risk}}, \\ 
1 - \alpha_0, & \textup{otherwise}.
\end{cases}
\end{align}
The function $f(\cdot)$ is a piecewise function, $\mathrm{risk}(\cdot)$ is the risk quantification function based on metrics such as TTC and PET, $\tau^{\mathrm{risk}}$ is a preset risk threshold, and ${\alpha_0}\in(0.5,1)$ is a constant value.
\end{definition}

\IncMargin{1em}
\begin{algorithm}
\SetKwData{TaskPolicy}{$\pi_{\theta}^{\text{task}}$}
\SetKwData{SafePolicy}{$\pi_{\phi}^{\text{safe}}$}
\SetKwData{ReplayBuffer}{$\mathcal{D}$}
\SetKwData{Priority}{$p(t)$}
\SetKwData{Transition}{$(s_t, a_t, r_t^{\text{task}}, r_t^{\text{safe}}, s_{t+1})$}
\SetKwData{Risk}{$\text{risk}(s_t)$}
\SetKwData{Weight}{$w_b$}
\SetKwData{AdvantageTask}{$\hat{A}^{\text{task}}$}
\SetKwData{AdvantageSafe}{$\hat{A}^{\text{safe}}$}
\SetKwData{Gamma}{$\rho$}
\SetKwFunction{Sample}{sample}
\SetKwFunction{Compute}{compute}
\SetKwFunction{Update}{update}
\SetKwFunction{Store}{store}
\SetKwFunction{Adjust}{adjust}
\SetKwFunction{Execute}{execute}
\SetKwInOut{Input}{input}\SetKwInOut{Output}{output}
\Input{Initial task policy parameters $\theta$, safety policy parameters $\phi$, replay buffer $\mathcal{D}$ with priority $p(t) \propto \eta(t)^\kappa$}
\Output{Updated policies $\pi_{\theta}^{\text{task}}$ and $\pi_{\phi}^{\text{safe}}$}

\BlankLine
\For{episode $= 1$ to $M$}{
    \emph{Reset environment:} $s_0 \sim \rho_0$\;
    \For{$t = 0$ to $T-1$}{
        \emph{Sample action from task policy:} $a_t^{\text{task}} \sim \pi_{\theta}^{\text{task}}(\cdot|s_t)$\;
        \emph{Sample action from safety policy:} $a_t^{\text{safe}} \sim \pi_{\phi}^{\text{safe}}(\cdot|s_t, a_t^{\text{task}})$\;
        \emph{Compute safety weighting factor:} 
         $\alpha_t = f(\tau_t), \quad \tau_t = \text{risk}(s_t)$\;
        \emph{Adjust action based on safety policy:} $a_t = a_t^{\text{task}} + \alpha_t(a_t^{\text{safe}} - a_t^{\text{task}})$\;
        \emph{Execute action and observe:} $r_t^{\text{task}}, r_t^{\text{safe}}, s_{t+1}$\;
        \emph{Update priority:} $p(t) \leftarrow \tau_t$\;
        \emph{Store transition:} $\mathcal{D} \leftarrow (s_t, a_t, r_t^{\text{task}}, r_t^{\text{safe}}, s_{t+1}, \tau_t, \alpha_t)$\;
    }
    \For{batch $b \sim \mathcal{D}$ with IS weights $w_b$}{
        \emph{Compute advantages:} $\hat{A}^{\text{task}}$, $\hat{A}^{\text{safe}}$ via GAE\;
        \emph{Update task policy parameters:} $\nabla_{\theta}\mathcal{L}_{\text{task}} = \mathbb{E}_b[w_b(1-\alpha_t)\hat{A}^{\text{task}}\nabla\log\pi_{\theta}^{\text{task}}]$\;
        \emph{Update safety policy parameters:} $\nabla_{\phi}\mathcal{L}_{\text{safe}} = \mathbb{E}_b[w_b\alpha_t\hat{A}^{\text{safe}}\nabla\log\pi_{\phi}^{\text{safe}}]$\;
        \emph{Project parameters:} $\theta \leftarrow \Pi_{\|\theta\| \leq C}(\theta)$, $\phi \leftarrow \Pi_{\|\phi\| \leq 0.27C}(\phi)$\;
    }
}
\caption{Dynamic Residual Safe Reinforcement Learning}
\label{algo_dr_safe_rl}
\end{algorithm}
\DecMargin{1em}

\cref{algo_dr_safe_rl} illustrates the calculating procedure of the DRS-RL. By employing a dual-policy collaborative mechanism, the framework overcomes the limitations of a single-policy approach. We introduce a safety weighting factor to implement loss weighting, thereby enabling adaptive reinforcement in safety-critical scenarios. The loss functions are defined as
\begin{align}
\left\{ 
\begin{array}{*{20}{l}}
  {{L^{{\text{task}}}} =  - {J^{{\text{task}}}}(1 - \alpha )} \\ 
  {{L^{{\text{safe}}}} =  - {J^{{\text{safe}}}}\alpha } 
\end{array} 
\right..
\end{align}

The algorithm intuitively embodies the core concept of the weak-to-strong safety correction paradigm, which aims to deliver safety supervision at minimal parameter overhead.  Leveraging a dynamic adjustment mechanism, it adaptively modulates the intensity of safety corrections based on real-time risk assessments. This paradigm pioneers the integration of the weak-to-strong theory \cite{burns2023weak} into the multi-agent decision-making domain, thereby addressing the inherent trade-off between safety and performance.

To further substantiate the theoretical robustness of the hybrid policy, it is imperative to ensure that incorporating the safety correction term does not compromise policy convergence. The derived convergence theorem is presented as \cref{thm:safety_residual_convergence}.

\begin{theorem}[Safety Residual Convergence]  
\label{thm:safety_residual_convergence}
If the following conditions are satisfied:

\begin{enumerate}
\item $\pi^{\mathrm{task}}$ is $\beta$-Lipschitz continuous: 
\begin{align}
\left\| {{\pi ^{{\mathrm{task}}}}(s) - {\pi ^{{\mathrm{task}}}}(s')} \right\| \leqslant \beta \left\| {s - s'} \right\|.
\end{align}

\item The safety correction term is bounded:
\begin{align}
\left\| {{\pi ^{{\mathrm{safe}}}}(s) - {\pi ^{{\mathrm{task}}}}(s)} \right\| \leqslant \gamma \left\| {{\nabla _s}{\pi ^{{\mathrm{task}}}}(s)} \right\|.
\end{align}
\end{enumerate}

Then there exists a Lyapunov function $V(s) = {\mathbb{E}}[{Q^{{\mathrm{safe}}}}(s,a)] + \lambda D_{\mathrm{KL}}({\pi ^{{\mathrm{hybrid}}}}\parallel {\pi ^{{\mathrm{task}}}})$, such that
\begin{align}
\Delta V(s) \leqslant  - \eta \left( {\alpha (s){\text{h}}(s) + (1 - \alpha (s)){{\left\| {{\nabla _s}{\pi ^{{\mathrm{task}}}}} \right\|}^2}} \right).
\end{align}
\end{theorem}

This conclusion demonstrates the asymptotic stability of the DRS-RL framework. It further indicates that the residual safety correction does not impair task policy convergence, instead, it facilitates a seamless integration of safety and performance through adjustment of the safety weighting factor.

\subsection{Risk-Aware Prioritized Experience Replay}

Conventional temporal-difference error–based prioritized sampling methods result in the undersampling of safety-critical segments, which undermines the model's capacity to manage high-risk events.  To address this issue, we propose a risk-aware PER method grounded in the MADCZ model.  The core concept is establishing a mapping between risk intensity and sampling probability, thereby actively increasing the sampling frequency of safety-critical samples. 

The risk-aware PER method integrates the multi-dimensional risk metrics proposed by the MADCZ model to quantify scenario risks, the main indicators are as follows
\begin{align}
\left\{ 
\begin{aligned}
  &{\text{TT}}{{\text{C}}_{{\text{norm}}}} = {f_1}({\text{TTC}}) \\ 
  &{\text{PE}}{{\text{T}}_{{\text{norm}}}} = {f_1}({\text{PET}}) \\ 
  &{\mathbb{I}_{{\text{event}}}} = {f_2}({\text{s}}) 
\end{aligned} 
\right.,
\end{align}
where ${\text{TTC}_{{\text{norm}}}}\in(0,1]$ and ${\text{PET}_{{\text{norm}}}}\in(0,1]$ represent the normalized TTC and PET, respectively. The function ${f_1}\left(\cdot \right)$ employs  the reciprocal of $\text{TTC}$ and $\text{PET}$ to quantify temporal urgency.  The function ${f_2}\left(\cdot \right)$ serves as an indicator for dangerous events, returning 1 when such events occur and 0 otherwise.  A weighted fusion mechanism is utilized to construct the scenario risk quantification function
\begin{align}
{\text{risk(s)}} = {\lambda _1}{\text{TT}}{{\text{C}}_{{\text{norm}}}} + {\lambda _2}{\text{PE}}{{\text{T}}_{{\text{norm}}}} + {\lambda _3}\sum\limits_{k = 1}^K {{\beta _k}{\mathbb{I}_{{\text{event}}}}},
\end{align}
where $\lambda_1$, $\lambda_2$, and $\lambda_3$ denote weight coefficients, and $\beta_k$ represents the weight of the $k$-th event, determined by its urgency. ${\text{risk}}(s)\in[{w_0},1]$ is the risk quantification function, with $w_0$ serving as the preset minimum risk threshold to ensure baseline learning utility in routine scenarios. 

During the experience replay phase, the sampling probability $p_i$ of each sample is determined by the proportion of the risk value
\begin{align}
{p_i} = \frac{{{\text{ris}}{{\text{k}}_i}}}{{\sum\nolimits_{j = 1}^N {{\text{ris}}{{\text{k}}_j}} }}.
\end{align}

\section{Experiments}

In this section, we construct a MASCS set and perform comparative experiments on multiple algorithms. The experimental results are analyzed from both the training process and the testing results to evaluate the performance of DRS-RL.

\subsection{Multi-Agent Safety-Critical Scenario Set}

We constructed the MASCS set based on the autonomous driving validation benchmark dataset, Bench2Drive \cite{jia2025bench2drive}, which encompasses  44 types of interaction scenarios, including cut-in, overtaking, detour, and other driving conditions. To meet the demands of safety-critical scenarios, we developed a multi-dimensional evaluation system that incorporates conflict type classification, risk level quantification, and additional assessments to systematically characterize each scenario’s features.  Following algorithmic screening and human evaluation, we identified four scenarios—leading vehicle emergency braking (LVEB), occluded pedestrian intrusion (OPI), roadside parking cut-in (RPC), and intersection jaywalking (IJ)—with the following selection criteria: 

\begin{enumerate}[label=(\alph*)]
\item	High Collision Risk: The scenario exhibits a short TTC, necessitating swift responses.
\item	Dynamic Interaction: The scenario involves dynamic participants and encompasses both vehicle-to-vehicle and vehicle-to-pedestrian interactions.
\item	Scenario Coverage: In accordance with SAE J3016 guidelines \cite{SAEJ3016}, the selected scenarios address key test items, including automatic emergency braking (AEB) and vulnerable road user (VRU) protection. 
\item	Scenario Diversity: The scenarios span two typical environments—urban roads and intersections. 
\end{enumerate}

\begin{figure}[bt!]
    \centering
    \includegraphics[width=1\linewidth]{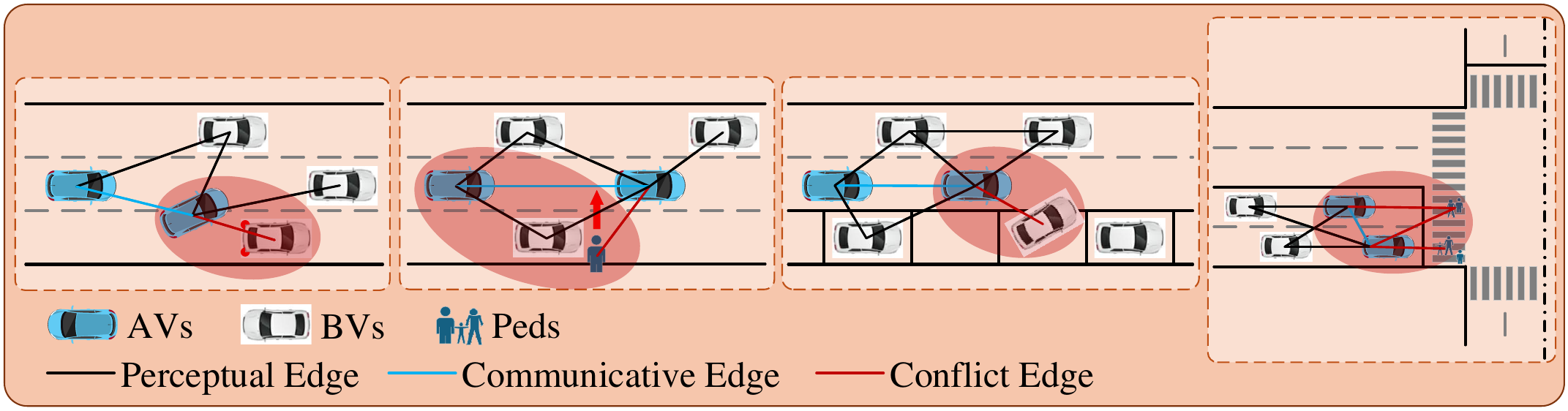}
    \caption{Multi-agent safety-critical scenario set. The first scenario is LVEB, where the AV should perform emergency braking or collaborate with the rear-side vehicle. The second scenario is OPI, where a pedestrian enters the lane from the blind spot. The third scenario is RPC, where AVs are traveling along a lane adjacent to a roadside parking lot, and a parked vehicle suddenly cuts in. The fourth scenario is IJ, where a group of pedestrians crosses the road at the intersection, significantly increasing the collision risk. Additionally, each scenario incorporates random variable parameters (including obstacle position, pedestrian triggering conditions, etc.) to create various variant scenarios.}
    \label{fig: scenario set}
\end{figure}

To address the limitation of single-agent scenario datasets that lack multi-agent interactions, we increased the number of AVs and elevated scenario complexity.  As shown in \cref{fig: scenario set}, we developed a MASCS set on the SUMO simulation platform \cite{sumo}, incorporating heterogeneous traffic participants—AVs, BVs, and Peds.  The scenario set encompasses a range of complex interaction challenges, including vehicle dynamic response, blind spot risk awareness, cut-in conflict resolution, and pedestrian group avoidance. This scenario set serves to validate the robustness of algorithms under safety-critical conditions while offering a standardized testing environment for autonomous driving decision-making systems. 


\subsection{Reward Function}

To address multi-objective decision-making in safety-critical scenarios, we developed a dual-reward optimization mechanism that integrates task-oriented and safety-protective rewards.  This mechanism guarantees baseline driving performance via task rewards while employing safety rewards to proactively mitigate hazardous scenarios. 

The task reward function is designed based on \cite{gao2024ethical} and \cite{liu2024mv}, covering sub-reward functions for basic safety, efficiency, and comfort
\begin{align}
&\left\{
\begin{array}{l}
  R_{\text{col}} = -C_0, \text{if collision occurs} \\ 
  R_{\text{dis}} = f_3\left( d \right) \\ 
  R_{\text{vel}} = f_4\left( \Delta v \right) \\ 
  R_{\text{com}} = f_5(a_{\text{lat}}, a_{\text{lon}})
\end{array}
\right.,
\end{align}
where $C_0$ denotes a constant collision penalty,  and $d$ represents the distance between the vehicle and its leading traffic participant. Function ${f_3}(\cdot)$ imposes a penalty on excessively short following distances to prevent rear-end collisions.  Function ${f_4}(\cdot)$ calculates rewards based on speed change $\Delta v$, whereas function ${f_5}(\cdot)$ quantitatively assesses comfort using lateral acceleration $a_{\text{lat}}$ and longitudinal acceleration $a_{\text{lon}}$.

The safety reward function is derived from the task reward function and further integrates advanced safety objectives.  It incorporates risk-sensitive reward components to augment risk-avoidance capabilities in safety-critical scenarios.  For the LVEB, OPI, and RPC scenarios, avoidance rewards are employed, whereas the IJ scenario utilizes a braking reward, 
\begin{align}
&\left\{
\begin{array}{l}
  R_{\text{avoid}} = e^{ - \frac{d}{\lambda }} (v - v_f) \min\left( \frac{|\delta|}{\delta_0}, 1 \right) \\ 
  R_{\text{brake}} = e^{ - \frac{d}{\lambda }} \frac{1}{\text{TTC} + \epsilon} \min\left( \frac{|a|}{a_0}, 1 \right)
\end{array}
\right.,
\end{align}
where $\lambda$ denotes the distance decay coefficient, $v_f$ represents the speed of the leading obstacle, $\delta_0$ is the avoidance angle derived from the obstacle's geometric characteristics, and $a_0$ is the desired deceleration.

\subsection{Results}

We conducted simulation experiments on the proposed DRS-RL framework based on the proximal policy optimization algorithm \cite{schulman2017proximal} (DRS-PPO), and compared its performance against traditional centralized DQN \cite{mnih2013playing} (CDQN), centralized double dueling DQN \cite{wang2016dueling} (CD3QN), and centralized PPO (CPPO). 

\subsubsection{Training Results}
\begin{figure*}[bt!]
    \centering
    \subfigure[LVEB.]{
        \begin{minipage}[t]{0.24\linewidth}
            \includegraphics[width=4.55cm]{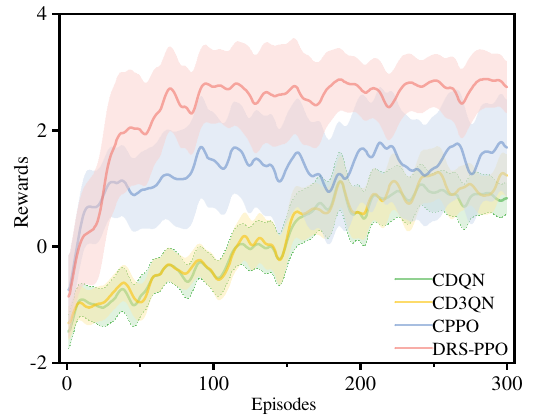}
            
        \end{minipage}
    } 
    \subfigure[OPI.]{
        \begin{minipage}[t]{0.225\linewidth}
            \includegraphics[width=4.35cm]{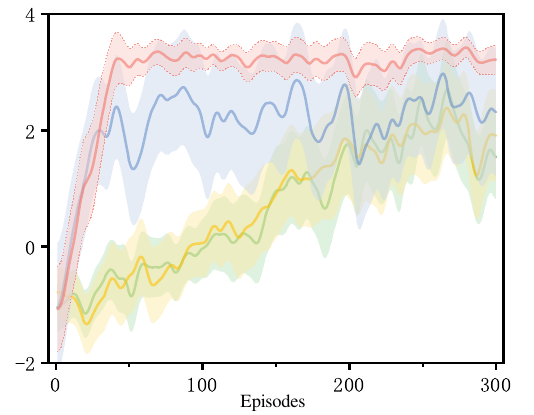}
            
        \end{minipage}
    }
    \subfigure[RPC.]{
        \begin{minipage}[t]{0.225\linewidth}
            \includegraphics[width=4.35cm]{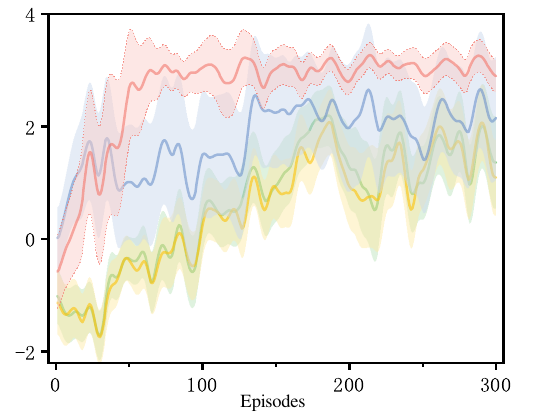}
            
        \end{minipage}
    }
    \subfigure[IJ.]{
        \begin{minipage}[t]{0.225\linewidth}
            \includegraphics[width=4.35cm]{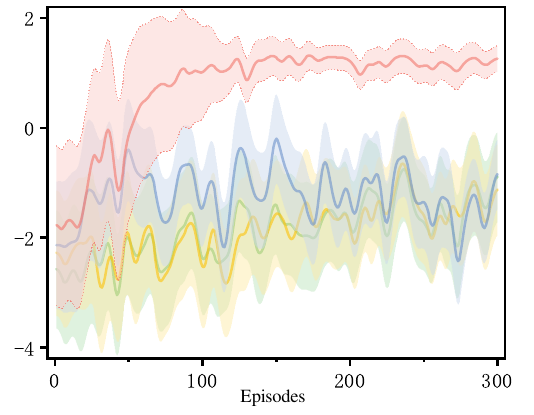}
        \end{minipage}
    }
    \caption{Reward curves of the four multi-agent safety-critical scenarios. The shaded areas show the standard deviation for 5 random seeds.}
    \label{fig:exp1}
\end{figure*}

To accurately evaluate the algorithm's overall performance,  we employ normalized reward \cite{liu2024mv} as a metric for training effectiveness. The reward curves are shown in \cref{fig:exp1}. Notably, all four algorithms employ graph convolutional networks (GCN) \cite{kipf2016semi} to process the topological structure information in the scenarios. The experimental results demonstrate that policy-based algorithms generally outperform value-based approaches.  Ablation experiment results reveal that the reward curve of the DRS-PPO algorithm surpasses that of the conventional CPPO algorithm, reflecting faster convergence and reduced variance. These results not only demonstrate that the proposed method significantly enhances the performance of the PPO algorithm in MASCS, but also confirm the reliability of \cref{thm:safety_residual_convergence}.

\subsubsection{Numerical Testing Results}
We tested and evaluated the trained models and the main indicators of the testing experiments are shown in \cref{tab:performance_comparison}. In the LVEB scenario, the DRS-PPO method achieved the highest average speed and zero collision rate. The average lateral and longitudinal accelerations were kept within a comfortable range. It should be noted that other algorithms had higher collision rates, leading to premature simulation termination and consequently shorter travel times. In the OPI and RPC scenarios, DRS-PPO similarly exhibited the lowest collision rates and higher average speeds.  In the IJ scenario, traditional methods experienced catastrophic failures, whereas the DRS-PPO algorithm achieved a collision rate of 4\%, thereby enhancing safety by approximately 92.17\%.  This risk avoidance was attained through active speed reduction (average speed of 7.13 m/s), thereby validating the efficacy of the safety correction mechanism.

\begin{table}[htbp]
\centering
\caption{Performance comparison across scenarios and metrics.}
\label{tab:performance_comparison}
\begin{tabular}{lccccc}
\toprule
\textbf{Scenarios} & \textbf{Metric} & \textbf{CDQN} & \textbf{CD3QN} & \textbf{CPPO} & \textbf{DRS-PPO} \\
\midrule
\multicolumn{1}{c}{\multirow{6}{*}{LVEB}} 
& CR & 48.00 & 33.00 & 31.50 & \textbf{0.00} \\
& AS & 13.80 & 13.79 & 13.93 & \textbf{14.05} \\
& TT & \textbf{10.19} & 10.82 & 10.93 & 11.35 \\
& ALA & 0.04 & 0.05 & \textbf{0.03} & 0.10 \\
& ALO & 1.85 & 1.80 & \textbf{1.79} & 1.82 \\
& Reward & 1.32 & 1.78 & 1.69 & \textbf{3.21} \\
\midrule
\multicolumn{1}{c}{\multirow{6}{*}{OPI}} 
& CR & 21.00 & 32.00 & 6.00 & \textbf{1.00} \\
& AS & 13.99 & 14.04 & \textbf{14.27} & 14.16 \\
& TT & 9.56 & \textbf{9.10} & 9.90 & 10.35 \\
& ALA & \textbf{0.04} & \textbf{0.04} & 0.06 & 0.06 \\
& ALO & \textbf{1.83} & 1.87 & 1.87 & 1.85 \\
& Reward & 2.30 & 1.85 & 3.03 & \textbf{3.23} \\
\midrule
\multicolumn{1}{c}{\multirow{6}{*}{RPC}} 
& CR & 19.50 & 33.00 & 7.00 & \textbf{1.00} \\
& AS & 13.27 & 12.66 & 13.03 & \textbf{13.86} \\
& TT & 10.79 & \textbf{10.52} & 11.71 & 10.91 \\
& ALA & \textbf{0.08} & \textbf{0.08} & 0.10 & 0.13 \\
& ALO & 1.78 & 1.74 & \textbf{1.62} & 1.83 \\
& Reward & 2.09 & 1.61 & 2.38 & \textbf{3.20} \\
\midrule
\multicolumn{1}{c}{\multirow{6}{*}{IJ}} 
& CR & 97.50 & 93.50 & 97.50 & \textbf{4.00} \\
& AS & 9.60 & \textbf{9.80} & 9.34 & 7.13 \\
& TT & \textbf{5.73} & 6.09 & 5.77 & 16.15 \\
& ALA & 0.17 & 0.41 & 0.17 & \textbf{0.00} \\
& ALO & 1.50 & 1.11 & 1.32 & \textbf{0.91} \\
& Reward & -0.45 & -0.32 & -0.58 & \textbf{1.85} \\
\bottomrule
\end{tabular}
\begin{tablenotes}
\item CR: Collision rate (\%), AS: Average speed (${\rm{m/s}}$), TT: Travel time ($\mathrm{s}$), ALA: Avg. lateral acceleration (${\rm{m/s^{2}}}$), ALO: Avg. longitudinal acceleration (${\rm{m/s^{2}}}$).
\end{tablenotes}
\end{table}

In summary, DRS-PPO consistently achieved the highest reward metrics across all scenarios, exhibiting an average improvement of 67.38\% relative to the next-best algorithm. It attains high safety while concurrently ensuring traffic efficiency and passenger comfort. The DRS-RL framework effectively addressed several challenges, including inadequate lightweight design of security models, performance limitations under safety constraints, difficulties in modeling the DCZ, and biases in data distribution.

\subsubsection{Analysis of Trajectories}

\cref{fig:trajs} presents examples of the AVs' trajectories during testing (a running video is available in the appendix).  It illustrates how AVs make a series of decisions and eventually complete the driving task safely. In the first three scenarios, AVs trained with the DRS-PPO algorithm learned to change lanes to avoid conflicting obstacles ahead. In the intersection scenario, they also learned braking strategies to yield to pedestrians. 

            
            
            

\begin{figure*}[bt!]
    \centering
    \subfigure[LVEB.]{
        \begin{minipage}[t]{0.23\linewidth}
            \includegraphics[height=2.7cm]{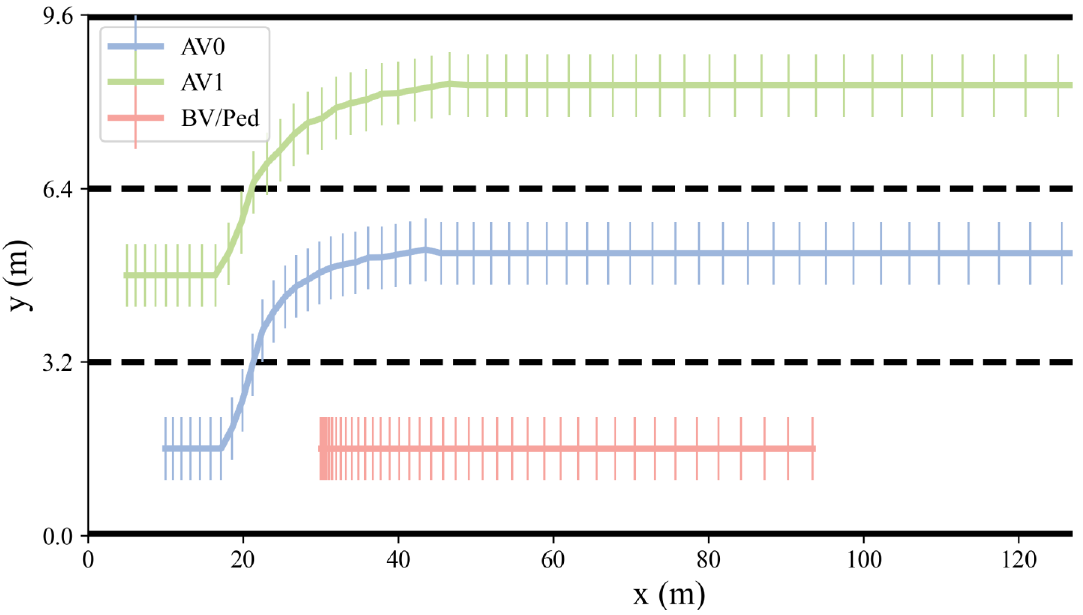}
            
        \end{minipage}
    } 
    \subfigure[OPI.]{
        \begin{minipage}[t]{0.23\linewidth}
            \includegraphics[height=2.7cm]{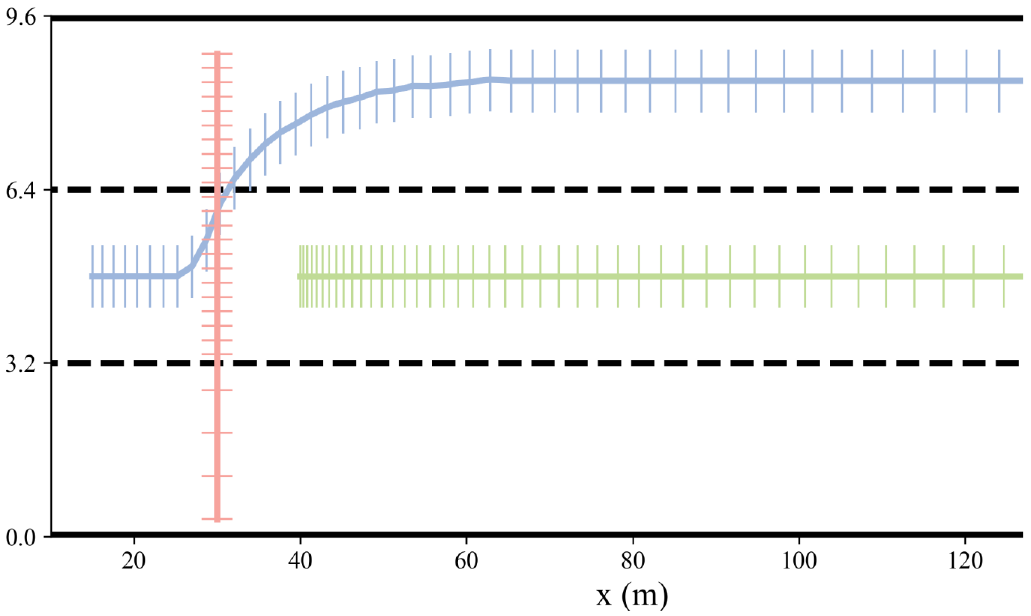}
            
        \end{minipage}
    }
    \subfigure[RPC.]{
        \begin{minipage}[t]{0.23\linewidth}
            \includegraphics[height=2.7cm]{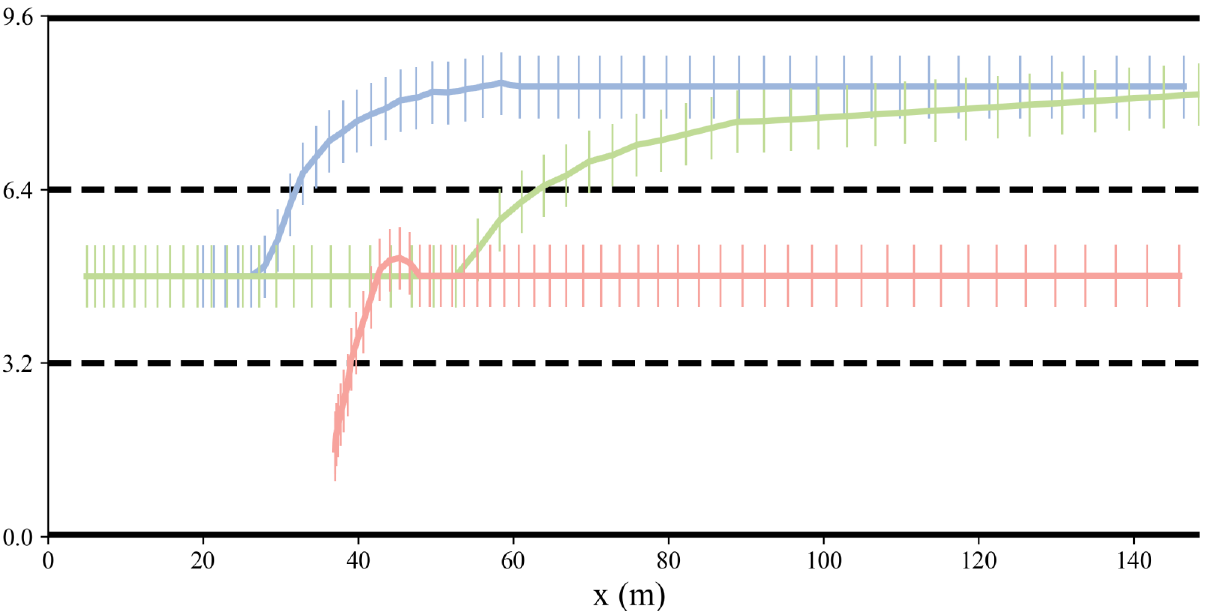}
            
        \end{minipage}
    }
    \subfigure[IJ.]{
        \begin{minipage}[t]{0.23\linewidth}
            \includegraphics[height=2.7cm]{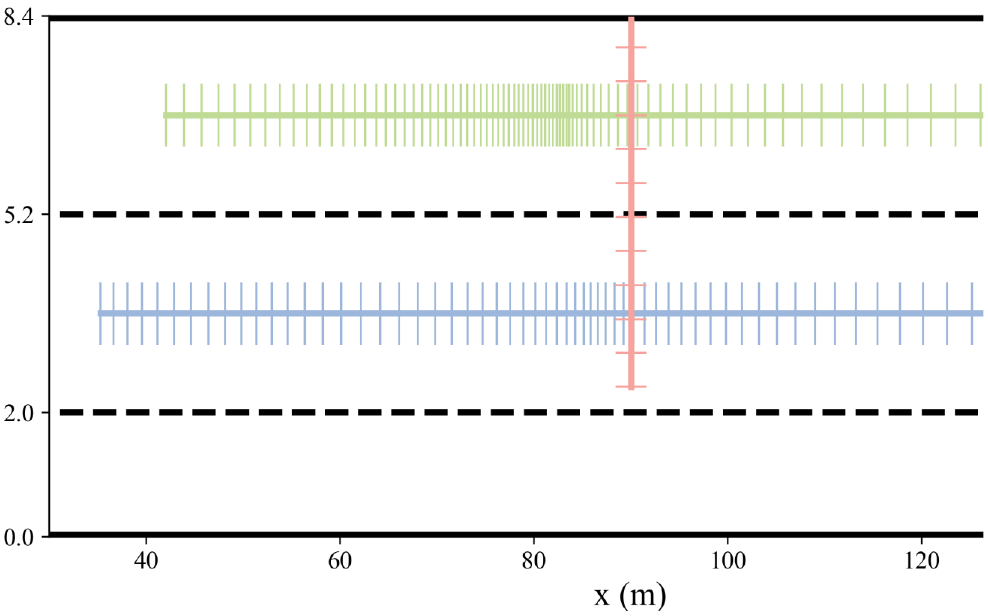}
        \end{minipage}
    }
    \caption{Trajectories in the four multi-agent safety-critical scenarios showed the learned policy.}
    \label{fig:trajs}
\end{figure*}

\section{CONCLUSIONS}

To resolve the decision-making challenges in MASCS, we propose a MADCZ modeling method to precisely quantify spatiotemporal coupling risks.  Furthermore, the DRS-RL framework and risk-aware PER method are presented, marking the inaugural application of weak-to-strong theory in multi-agent decision-making.  This approach dynamically balances safety and performance while mitigating data bias. Experiments on the constructed MASCS set demonstrate that our method significantly outperforms traditional RL algorithms in terms of collision rate, average speed, and comfort. Specifically, the collision rate is reduced by up to 92.17\%, validating its safety effectiveness in MASCS.

In future research, we intend to increase the number of AVs and develop a more diverse set of safety-critical scenarios.  Furthermore, we will further refine existing methods in extreme emergency scenarios to facilitate the reliable deployment of autonomous driving systems.



\bibliographystyle{ieeetr}
\small\bibliography{reference}

\end{document}